%% file: 2024-silk.tex
\pdfoutput=1

\documentclass[11pt]{article}
\usepackage[dvipsnames]{xcolor}

\usepackage[]{acl}

\usepackage{times}
\usepackage{latexsym}

\usepackage[T1]{fontenc}

\usepackage[utf8]{inputenc}

\usepackage{microtype}

\usepackage{inconsolata}

\usepackage{amsmath}

\usepackage{booktabs}
\usepackage{multirow}
\usepackage{enumitem}
\usepackage{makecell}
\usepackage{diagbox}
\usepackage{fontawesome5}

\usepackage{graphicx}
\usepackage{tikz}

\newcommand{\silk}{\texttt{silk}}
\definecolor{sne-red}{HTML}{EE2967}
\definecolor{sne-yellow}{HTML}{ffd966}
\definecolor{sne-green}{HTML}{008B72}
\definecolor{sne-blue}{HTML}{613F99}
\definecolor{modif}{HTML}{000000}

\newlength{\myl}
\newcommand{\sign}[1]{\textbf{#1}$^{\dagger}$\hspace{-\myl}}
\newcommand{\ssign}[1]{#1$^{\dagger}$\hspace{-\myl}}

\title{Unsupervised Domain Adaptation for Keyphrase Generation using Citation Contexts}

\author{Florian Boudin \\
  JFLI, CNRS, Nantes University, France \\
  \texttt{florian.boudin@univ-nantes.fr} \\\And
  Akiko Aizawa \\
  National Institute of Informatics, Japan \\
  \texttt{aizawa@nii.ac.jp} \\}

\begin{document}
\maketitle
\begin{abstract}
Adapting keyphrase generation models to new domains typically involves few-shot fine-tuning with in-domain labeled data.
However, annotating documents with keyphrases is often prohibitively expensive and impractical, requiring expert annotators.
This paper presents \silk{}, an unsupervised method designed to address this issue by extracting silver-standard keyphrases from citation contexts to create synthetic labeled data for domain adaptation.
Extensive experiments across three distinct domains demonstrate that our method yields high-quality synthetic samples, resulting in significant and consistent improvements in in-domain performance over strong baselines.
\end{abstract}

\section{Introduction}

Keyphrase generation aims at automatically predicting a set of keyphrases ---words or phrases that represent the main concepts--- given a source text.
Because they distill the important information from documents, keyphrases are useful for many applications in natural language processing and information retrieval, most notably for document indexing~\cite{10.1145/42005.42016,zhai-1997-fast,10.1145/312624.312671,GUTWIN199981,boudin-etal-2020-keyphrase} and summarization~\cite{10.1145/564376.564398,wan-etal-2007-towards,10.1145/3442381.3449906,koto-etal-2022-lipkey}.
Keyphrase generation differs from its extractive counterpart in that it requires the capability of predicting keyphrases that do not necessarily appear
in the source text~\cite{liu-etal-2011-automatic,meng-etal-2017-deep}.
Current models for this task are built upon sequence-to-sequence models, and achieve remarkable prediction performance when a large amount of labeled data is available~\cite{meng-etal-2021-empirical}.

However, keyphrase-labeled data is notably scarce even for resource-rich languages.
To date, there are only a handful of available datasets large enough to train keyphrase generation models, therefore restricting their applicability to specific domains~\cite{ye-wang-2018-semi,wu-etal-2022-representation,garg-etal-2023-data}.
Here, we are concerned with generating keyphrases from scientific papers, for which datasets only exist in the broader scope of computer science~\cite{meng-etal-2017-deep,mahata2022ldkp} and biomedicine~\cite{houbre-etal-2022-large}. 
This data scarcity issue is all the more important since current models demonstrate very limited generalization capabilities~\cite{gallina-etal-2019-kptimes,10.1145/3383583.3398517,meng-etal-2021-empirical}.
All of this, coupled with the high computational cost of training models, underscores the necessity of developing domain adaptation methods for keyphrase generation.

\begin{figure}[!t]
    \centering
    \resizebox{\linewidth}{!}{
    \includegraphics{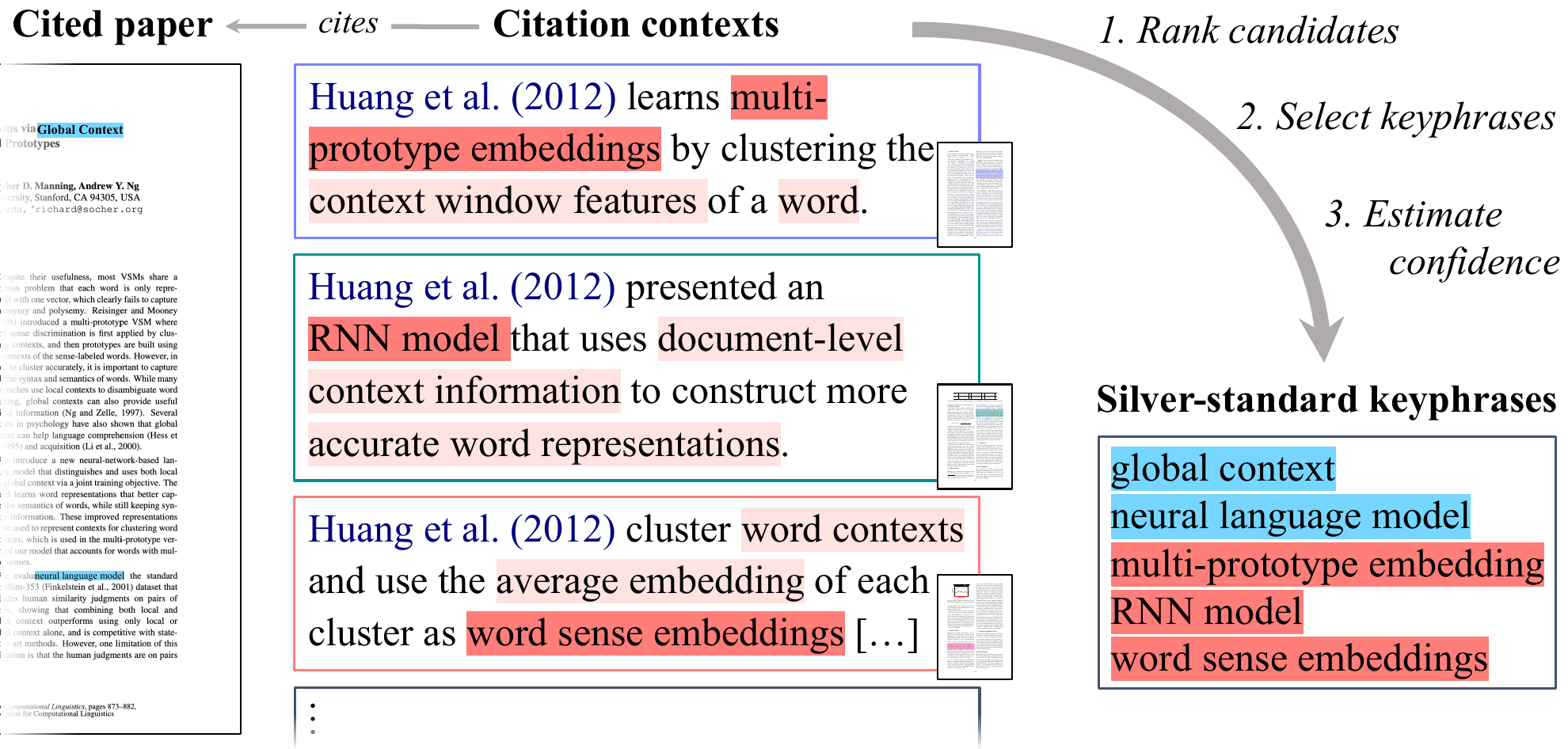}
    }
    \caption{Illustration of the \silk{} method for mining silver-standard keyphrases (highlighted in red) from citation contexts and generating synthetic samples for adapting models to new domains.}
    \label{fig:cc}
    \vspace{-1em}
\end{figure}

An effective strategy for addressing this challenge involves low-resource fine-tuning~\cite{wu-etal-2022-representation,meng-etal-2023-general}, wherein a pre-trained model is exposed to a limited amount of in-domain data with annotated keyphrases.
Nevertheless, annotating even a limited number of documents can be prohibitively expensive, and often impractical due to the necessity for expert annotators~\cite{chau-etal-2020-understanding}.
Finding a way to collect such data in an unsupervised fashion would open up possibilities for effortlessly adapting models to new domains.
Here, we propose \silk{}, a method to do so that relies on extracting \underline{sil}ver-standard \underline{k}eyphrases from citation contexts to generate synthetic labeled data for domain adaptation (see Figure~\ref{fig:cc}).

Citation contexts ---text passages within the citing document containing the reference--- often highlight the contributions of a cited paper, and have be shown to be useful not only for paper summarization~\cite[\textit{inter alia}]{nakov2004citances,schwartz-hearst-2006-summarizing,mei-zhai-2008-generating,abu-jbara-radev-2011-coherent,mao-etal-2022-citesum}, but also for tasks such as claim verification~\cite{wadden-etal-2020-fact} or information extraction~\cite{viswanathan-etal-2021-citationie}.
In this paper, we advocate for using citation contexts, specifically in the \emph{mining of phrases representing the key concepts of cited papers}, to generate synthetic data for adapting keyphrase generation models to new domains.
Earlier research on keyphrase extraction has emphasized the value of citation context information as a feature for ranking phrases~\cite{das_gollapalli2014extracting,caragea-etal-2014-citation}.
We take this idea further and explore how it can be applied to create silver-labeled in-domain data for fine-tuning keyphrase generation models.
Our contributions can be summarized as follows:

\begin{enumerate}[label=(\textbf{\arabic*})]
    \item We propose \silk{}, a method that leverages citation contexts to create synthetic samples of documents paired with silver-standard keyphrases for adapting keyphrase generation models to new domains.

    \item We apply our method on three distinct scientific domains ---namely, Natural Language Processing, Astrophysics and Paleontology---, thereby creating new adaptation data for each domain.
    We further provide three human-labeled test sets to assess the performance of keyphrase generation across these domains.
    We view this effort as a significant contribution of our work.

    \item We conduct experiments on few-shot fine-tuning a pre-trained model for keyphrase generation and report significant improvements in in-domain performance using synthetic samples generated by \silk{}.
    Additionally, we undertake further experiments to validate the quality of the synthetic samples through both empirical (\S\ref{subsec:confidence-ranking}) and human (\S\ref{subsec:manual-evaluation}) evaluations, and we examine whether our adapted models experience catastrophic forgetting of the initial domain (\S\ref{subsec:forgetting}) \textcolor{modif}{or exhibit bias towards keyphrases from highly cited papers (\S\ref{subsec:bias})}.

\end{enumerate}

Our code, model weights and data are available at \url{https://github.com/boudinfl/silk/}.

\section{Method}
\label{sec:method}

This section describes the implementation details of our method for producing synthetic fine-tuning data from citation contexts.
Given a collection of in-domain scientific documents $\mathcal{D}$, we start by extracting the subset of sentences that contain citation anchors to build a set of citation contexts~$\mathcal{C}$.
Heuristics are applied to filter out citation contexts that either reference a document $d \not\in \mathcal{D}$ or whose purpose of citing is ambiguous (i.e.~containing multiple scattered citation anchors throughout the text).
For each cited document $d \in \mathcal{D}$, we extract all the phrases\footnote{We use spacy (\texttt{en\_core\_web\_sm} model) and consider the noun phrases (\texttt{/Adj*Noun+/}) in their lemma forms as candidates. Irrelevant candidates are filtered out using a stoplist of high-frequency phrases.} from its title $t_d$, abstract $a_d$ and corresponding citation contexts $c_d$ to build a set of silver-standard keyphrase candidates $\mathcal{P}_d$.
Our method for generating synthetic samples from pairs of $(d,~\mathcal{P}_d)$ involves three steps, which are described below.

\subsection*{Step 1: Ranking Keyphrase Candidates}
\label{subsec:step-1}

We rank each keyphrase candidate $p \in \mathcal{P}_d$ based on three criteria: 
\begin{itemize} 
    \item its \textcolor{WildStrawberry}{salience}, defined as the presence of $p$ in $t_d$, $a_d$ and $c_d$.
    Here, we assume that a phrase simultaneously occurring in all elements holds greater importance that a phrase found solely in one or two of them.
    A boosting parameter $\alpha = \{1, 1.5, 2\}$ is introduced to prioritize phrases based on the number of elements in which they appear.

    \item its \textcolor{Emerald}{relevance}, computed as the cosine distance between the embedding vectors of $p$ and $t_d$.
    We use the title as a high-level summary of the document, and assume that relevant phrases should be semantically close to it.
    We leverage SPECTER\footnote{\url{https://huggingface.co/sentence-transformers/allenai-specter}}~\cite{cohan-etal-2020-specter}, a BERT-based model pre-trained on scientific documents, to compute the embedding vectors.

    \item its \textcolor{Periwinkle}{reliability}, estimated by the number of citation contexts in which $p$ occurs.
    We rely on the citation context frequency as a means to estimate how reliable a phrase is, the hypothesis being that phrases that appears in multiple citation contexts are more likely to be reliable.
    Specifically, we use the log-frequency of $p$ in $c_d$ to squash the range of values in a log-scale. 
    
\end{itemize}

More formally, given a document $d$, the score of a keyphrase candidate $p$ is calculated as follows:
\begin{multline}
\mathrm{score}(p, d) =
{\color{WildStrawberry} \alpha_{(p)}}
\cdot
{\color{Emerald} \mathrm{cos}\textrm{-}\mathrm{sim}\bigl(\mathrm{emb}(p), \mathrm{emb}(t_d)\bigr)} \\
\cdot
{\color{Periwinkle}\log\bigl(\mathrm{freq}_{\mathrm{cc}}(p)\bigr)}
\label{eq:scoring}
\end{multline}
where $\mathrm{emb}(s)$ denotes the embedding vector output of SPECTER for input text $s$, $\mathrm{freq}_{\mathrm{cc}}(p)$ is the number of citation contexts in which $p$ occurs.

\subsection*{Step 2: Selecting Silver-Standard Keyphrases}
\label{subsec:step-2}

To select the optimal subset of phrases from $\mathcal{P}_d$, we define a set of constraints that mirrors the typical characteristics found in gold standard keyphrases of scientific papers.
Building on past observations, our objective is to select between 3 and 5 phrases per document, comprising up to 3 phrases from its content (i.e.~occurring in $t_d$ or $a_d$) and the remainder from the citation contexts.
We promote the selection of diverse keyphrases by introducing a maximum cross-phrase similarity threshold parameter $\lambda_x$.
This parameter prevents the inclusion of redundant candidates, as determined by the cosine distance between their embedding vectors.
Because candidates extracted from citation contexts are inherently noisy, we introduce a second threshold parameter $\lambda_r$ to filter out spurious candidates based on their \textcolor{Emerald}{relevance} scores.

\subsection*{Step 3: Ordering Samples by Confidence}
\label{subsec:step-3}

The final step involves ordering the cited documents based on how confident our method is in its silver-standard keyphrases, and selecting the top-$N$ ranked documents as synthetic labeled data.
Here, we determine the confidence of our method by averaging the scores of its silver-standard keyphrases, as computed in Equation~\ref{eq:scoring}.
We remind that our objective is to \emph{generate small, high quality in-domain data for fine-tuning keyphrase generation models}, which advocates for a conservative approach.

\section{Datasets}

We use the widely adopted KP20k dataset~\cite{meng-etal-2017-deep} as a starting point for pre-training keyphrase generation models.
This dataset contains $\approx514\mathrm{K}$ scientific documents (titles and abstracts) paired with author-assigned keyphrases in the broader domain of computer science.
We investigate the effectiveness of our domain adaptation method across three distinct scientific domains: Natural Language Processing (\texttt{nlp}), Astrophysics (\texttt{astro}), and Paleontology (\texttt{paleo}).
These domains differ with increasing distances from the initial KP20k dataset, with \texttt{nlp} being the closest and \texttt{paleo} standing as the furthest.
This section gives details about the data we use for each domain, presents the statistics of the resulting synthetic in-domain data we generate, and describes how we collect\footnote{Detailed information on the sources can be found in~\ref{subsec:appendic-venues-for-gold}.} annotated test data to validate the usefulness of our method for domain adaptation.
We set our method parameters (step 2 in \S\ref{subsec:step-2}) based on their observed values in the validation split of KP20k, specifically, $\lambda_x=0.85$ and $\lambda_r=0.75$.

\subsection{Natural Language Processing \hfill (\texttt{nlp})}

For the \texttt{nlp} domain, we use the ACL Anthology Sentence Corpus\footnote{\url{https://kmcs.nii.ac.jp/resource/AASC/}} that contains the sentences of 65\,662 papers from the ACL Anthology up until 2022.
For quality reasons, we only consider sentences from papers published in the last 20 years (2003 and upwards) and occurring within the introduction and related work sections.
From these, we extracted 260\,324 citation contexts with the restriction that they include at least one citation to a paper within the ACL Anthology.
For each cited paper, we applied our method to extract silver-standard keyphrases from citation contexts, resulting in a confidence-ordered list of 6\,199 synthetic samples.

As most papers in the ACL Anthology do not provide keyphrases, we mainly relied on NLP-related conferences and journals to compile the test data for the \texttt{nlp} domain.
More precisely, we manually collected a set of 212 documents (title and abstract) with author-assigned keyphrases from a variety of sources (e.g.~LREC, SIGIR, CIKM).

\subsection{Astrophysics \hfill (\texttt{astro})}

For the \texttt{astro} domain, we use the unarXive 2022 dataset~\cite{Saier2023unarXive} that contains 1.9M full-text papers from arXiv.
We selected the subset of 198\,349 papers that belong to the Astrophysics category (astro-ph), and extracted 133\,320 citation contexts originating from the introduction sections of these papers.
Applying our method for each cited paper produces in a confidence-ordered list of 2\,680 synthetic samples.

For the \texttt{astro} test data, we manually collected a set of 255 documents (title and abstract) paired with author-assigned\footnote{It should be noted that controlled vocabularies are also used to index papers in astrophysics, but these are not considered in our study.} keyphrases from both arXiv and journals.
To ensure topic diversity, we uniformly selected 20 documents from each astrophysics sub-category in arXiv and retrieved documents from broader-scope journals.

\subsection{Paleontology \hfill (\texttt{paleo})}

To the best of our knowledge, there is no dataset of scientific papers available for the \texttt{paleo} domain.
Thus, we collected 12\,353 open- or free-access papers in PDF format from a wide range of journals in Paleontology.\footnote{See Table~\ref{tab:paleo-sources} in Appendix~\ref{sec:appendix} for the detailed sources.}
We use GROBID\footnote{\url{https://github.com/kermitt2/grobid}} for extracting the full-text from PDF papers, detecting inline citations and parsing bibliography, as it was shown to outperform other freely available tools~\cite{10.1007/978-3-031-28032-0_31,rohatgi-etal-2023-acl}.
From the XML output of GROBID, we extracted 53\,133 citation contexts from the introductory parts of the papers (i.e.~``\textit{Introduction}'', ``\textit{Materials and Methods'}' and ``\textit{Geological Settings}'').
With such a small collection, applying our method yields too few synthetic samples.
To generate sufficient data for fine-tuning keyphrase generation models, we adjusted the threshold for candidate relevance (i.e.~$\lambda_r=0.75 \rightarrow 0.60$) and queried the Semantic Scholar API\footnote{\url{https://www.semanticscholar.org/}} to include cited papers not present in our collection.
These modifications resulted in our method generating a confidence-ordered list of 2\,806 synthetic samples. 

For the \texttt{paleo} test data, we manually collected a set of 244 documents, each paired with author-assigned keyphrases, sourced from approximately 10 different journals that encompass a wide spectrum of palaeontological topics (e.g.~palaeogeography, palaeoecology or stratigraphy).

\subsection{Statistics and Analysis}

In this section, our aim is to deepen our understanding of the characteristics of the datasets we use for each domain and to assess how the compiled test data aligns with existing test datasets.

Table~\ref{tab:stats} summarizes the statistics of the datasets for each domain we apply our method on.
There is a noticeable diversity in characteristics across the datasets, with \texttt{nlp} showing the highest citation rate per document and \texttt{paleo} the lowest.
We suspect there are two reasons for this.
First, papers within the \texttt{nlp} domain seem to garner higher average citations compared to papers in the other two domains.
Second, papers from \texttt{paleo} tend to cite works from both related domains (e.g.~Biology, Geology) and sources outside our collection of gathered papers.
Conversely, the average number of candidate keyphrases per document ---those found in the title, abstract, or citation contexts--- remains stable across the domains ($\approx$80 candidates).

\begin{table}[t!]
    \centering
    \resizebox{\linewidth}{!}{%
    \input{tables/data_statistics}
    }
    \caption{Statistics for the datasets and the top-1K synthetics samples generated by \silk{} for each domain.}
    \label{tab:stats}
\end{table}

Upon examining the synthetic fine-tuning data generated by our method (restricted to the top-1K), we observe that \texttt{nlp} documents are nearly half the length of those in the \texttt{paleo} domain, while \texttt{astro} documents fall in-between.
These differences in length directly impact the ratio of absent keyphrases\footnote{We follow the definition of~\cite{boudin-gallina-2021-redefining} and consider keyphrases that do not match contiguous sequences of (stemmed) words in the source document as absent.}, decreasing from 24\% to below 10\%.
These numbers further decrease when computed beyond the top-1K, as the number of citation contexts declines and, consequently, as the pool of absent keyphrase candidates reduces.
Constraints we introduced for selecting the optimal subset of phrases allow for an average of about 4 silver keyphrases per document, predominantly unigrams and bigrams, which is in line with both past observations and the test data we compiled (see Table~\ref{tab:test_data}).

To analyze the disparities between the domains we selected, and also how they depart from KP20k (initial domain) and from other existing test datasets for keyphrase generation, we compare the main statistics of their test splits in Table~\ref{tab:test_data}.
Here, we include three additional datasets, Inspec~\cite{hulth-2003-improved}, NUS~\cite{10.1007/978-3-540-77094-7_41} and SemEval-2010~\cite{kim-etal-2010-semeval}, that are composed of scientific abstracts in the computer science domain.
Together with KP20k, these are likely the most commonly-used datasets for evaluating keyphrase generation models.
Overall, we observe many similarities between KP20k and the test data we collected for each domain, whether in terms of the number of gold keyphrases ($\approx$5 per document), their average length  ($\approx$2 tokens) or the ratio of absent keyphrases ($\approx$40\%).
This suggests a uniform trend in author-assigned keyphrases across scientific domains, which should facilitate generalization for keyphrase generation models.
It should be noted that higher number of gold keyphrases in NUS, SemEval-2010 and Inspec stems from their distinct annotation processes, with the former two combining author- and reader-assigned keyphrases and the latter relying on professional indexers.
Comparing the sizes of our domain-specific test data with those of the test splits in existing datasets shows that they are on a similar scale.

\begin{table}[t]
    \centering
    \resizebox{.49\textwidth}{!}{%
    \input{tables/test_data}
    }
    \caption{Statistics for the test data we collected for each domain in comparison with the commonly used test sets for keyphrase generation.}
    \label{tab:test_data}
\end{table}

\begin{figure}[!ht]
    \centering
    \resizebox{\linewidth}{!}{
    \includegraphics{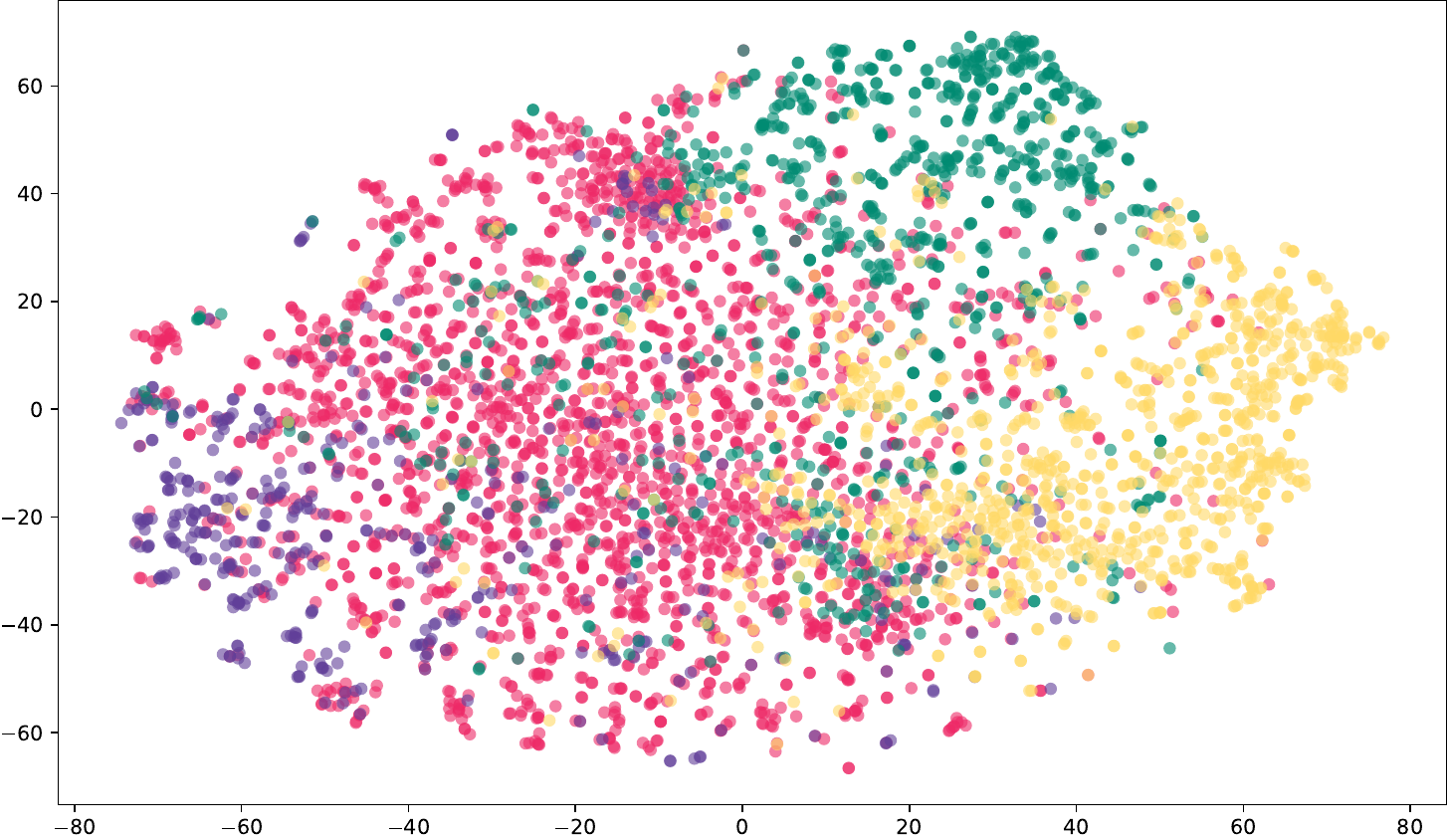}
    }
    \caption{t-SNE 2-D projections of the gold keyphrases from \textcolor{sne-red}{$\bullet$} KP20k, \textcolor{sne-blue}{$\bullet$} \texttt{nlp}, \textcolor{sne-green}{$\bullet$} \texttt{astro} and \textcolor{sne-yellow}{$\bullet$} \texttt{paleo}.
    We leverage SPECTER to compute the keyphrase embeddings and use the first 500 documents from KP20k for clarity.
    }
    \label{fig:t-sne}
\end{figure}

Lastly, we examine the differences between the domains from a semantic perspective.
Figure~\ref{fig:t-sne} shows a t-SNE visualization~\cite{JMLR:v9:vandermaaten08a} of the gold keyphrases in the test data that we collected for each domain and those of the KP20k test split.
We clearly discern the different domains within the vector space, roughly dividing it into four clusters. 
The most notable overlap occurs between \texttt{nlp} and KP20k (computer science), whereas \texttt{astro} and \texttt{paleo} exhibit clear separation.
These visual insights support our initial assumptions regarding the growing differences of our selected domains from KP20k, with \texttt{nlp} being the closest and \texttt{paleo} standing as the furthest.

\section{Experimental Settings}

\subsection{Initial Model}

We use BART~\cite{lewis-etal-2020-bart} as our initial pre-trained language model and perform fine-tuning on the KP20k training set for 15 epochs, following~\cite{wu-etal-2023-rethinking-model}.
BART was shown to yield state-of-the-art performance in keyphrase generation~\cite{zhao-etal-2022-keyphrase,wu-etal-2022-representation,meng-etal-2023-general}, surpassing other pre-trained language models, such as T5~\cite{wu-etal-2023-rethinking-model}.
Following previous work, we fine-tune BART in a \textsc{One2Many} setting~\citep{yuan-etal-2020-one}, that is, given a source text as input, the task is to generate keyphrases as a single sequence of delimiter-separated phrases.
During fine-tuning, gold keyphrases are arranged in the present-absent order which was found to give the best results~\cite{meng-etal-2021-empirical}.
At test time, we use either greedy decoding and let the model generate the most probable keyphrases, or beam search (K=20) and assemble the top-$k$ keyphrases from all the beams as the model output. 
Implementation details and training times are provided in Appendix~\ref{subsec:implementation-details}.

\subsection{Domain Adaptation}

For adapting our fine-tuned BART model to a specific domain, we continue fine-tuning it on the synthetic labeled data generated by \silk{} for 3 epochs.
Specifically, we use the top-$N$ most confident silver-labeled examples to further fine-tune BART, creating three gradually adapted models for each domain by varying $N \in \{500, 1K, 2K\}$.
We compare the effectiveness of our domain adaptation method with that of the only other unsupervised approach we are aware of, which is self-learning~\cite{ye-wang-2018-semi,meng-etal-2023-general}.
Self-learning consists in using a model to generate pseudo-labels for in-domain documents and then re-train itself on this data.
Here, we use our fine-tuned BART model to generate keyphrases for the same documents as those produced by \silk{}, and further fine-tune it on this self-labeled data for 3 epochs.

\subsection{Baselines}

Although the focus of this work is domain adaption, we also provide the results of several baselines as a point of reference.
The first baseline is MultiPartiteRank~\cite{boudin-2018-unsupervised}, an unsupervised method for keyphrase extraction that leverages graph-based ranking and topical information.
Despite being limited to present keyphrases, MultiPartiteRank yields the best results among non deep learning methods~\cite{do-etal-2023-unsupervised}.
We use the author's implementation provided by the \texttt{pke} toolkit~\cite{boudin-2016-pke}.\footnote{\url{https://github.com/boudinfl/pke}}
The second baseline is Yake~\cite{CAMPOS2020257}, another unsupervised method for keyphrase extraction that relies on statistical text features.
We use the author's implementation.\footnote{\url{https://github.com/LIAAD/yake}}
The third baseline is KeyBART~\cite{kulkarni-etal-2022-learning} in a zero-shot setting, a task-specific language model trained to learn rich representations of keyphrases.
We use the model weights released by the authors.\footnote{\url{https://huggingface.co/bloomberg/KeyBART}}
The fourth baseline is One2Set~\cite{ye-etal-2021-one2set}, a Transformer-based model that uses learned control codes to generate a set of keyphrases.
Trained on KP20k, this model achieves strong performance, often on-par with state-of-the-art models~\cite{wu-etal-2023-rethinking-model}.
We use the model weights released by the authors.\footnote{\url{https://github.com/jiacheng-ye/kg_one2set}}

\subsection{Datasets and Evaluation Metrics}

We use the test split of KP20k for evaluating the initial performance of the models, and our manually collected test sets to assess their in-domain performance.
Detailed statistics for these datasets are presented in Table~\ref{tab:test_data}.
Following common practice, we evaluate the performance of the models in terms of $F_1$ score using exact match between gold and predicted keyphrases.
Stemming (Porter stemmer) is applied to reduce the number of mismatches and duplicates are removed.
We compute the scores both at the top-$k$ predicted keyphrases with $k \in {5, 10}$, and at the number $M$ of keyphrases predicted by the models as proposed in~\cite{yuan-etal-2020-one}.
For $F_1@k$ scores, if the number of predicted keyphrases is below $k$, we append incorrect predictions until it reaches exactly $k$ keyphrases.
We also report scores for present and absent keyphrases separately to get more insights about the extractive and generative capabilities of the models.
We compute the Student's paired t-test to assess the statistical significance of our results at $p < 0.05$.

\subsection{Performance of Models on KP20k}

Table~\ref{tab:baseline-results} presents the results of our fine-tuned BART model (hereafter denoted as BART-FT) and the baselines on the test split of KP20k.
It should be noted that MultiPartiteRank and Yake cannot be assessed using $F_1@M$ as they require setting a top-$k$ parameter, and that One2Set cannot be assessed using $F_1@10$ since it only outputs the most probable keyphrases ($\approx$7 per document).
Overall, BART-FT demonstrates superior performance, significantly outperforming the baselines for both all and only the present keyphrases.
We observe that One2Set achieves the best scores for the absent keyphrases, confirming previous findings~\cite{wu-etal-2023-rethinking-model}.
In light of these results, we argue that BART-FT is a strong model for keyphrase generation, providing a solid basis for the application of our domain adaptation method.

\begin{table}[ht!]
    \centering
    \resizebox{\linewidth}{!}{
    \input{tables/baseline-results}
    }
    \caption{Performance comparison of our fine-tuned BART model and baseline models on the KP20k test set, with $\dagger$ indicating statistical significance. Scores for present and absent keyphrases separately are reported.}
    \label{tab:baseline-results}
\end{table}

\section{Results}

\begin{table*}[t]
    \centering
    \input{tables/adaptation-results.v4} 
    \caption{Performance of keyphrase generation models on the \texttt{nlp}, \texttt{astro} and \texttt{paleo} domains for all keyphrases (i.e. present and absent combined). Values in \textbf{bold} indicate best scores and $\dagger$ indicates significance over BART-FT.}
    \label{tab:adaptation-results}
\end{table*}

Table~\ref{tab:adaptation-results} presents the results of the keyphrase generation models and our domain adaptation method on each domain.\footnote{See Table~\ref{tab:pres-abs-results} in Appendix~\ref{sec:appendix} for present/absent results.}
We observe that \texttt{silk} brings consistent and significant improvements over BART-FT on the three domains.
The best overall performance is achieved by fine-tuning the model with the top-1K most confident synthetic samples, however gains are observed with just the top-500 samples.
Self-learning for domain adaptation yields only marginal gains at best and often degrades performance.
This suggests that the initial performance of BART-FT on these domains is not sufficient to generate high-quality pseudo-labels.
A closer look at the numbers shows that BART-FT performs comparably on \texttt{nlp} as it does on KP20k (see Table~\ref{tab:baseline-results}), but it gives substantially lower scores on \texttt{paleo} and \texttt{astro}.
This empirically confirms the growing distance between KP20k and these three domains, correlating model performance with the distance from the initial domain.

Among the three domains, \texttt{paleo} poses the greatest challenge for our method.
We see two main reasons for this.
First, the limited size of our collection of full-text papers ($\approx$12K), and the necessary parameter adjustments made to accommodate it, adversely affect the quality of the synthetic samples.
Second, the \texttt{paleo} domain in itself appears to be more challenging to handle due to its interdisciplinary nature, spanning subjects such as geology, biology, history, and ecology, among others.
Examining the performance of the baselines, we observe the poor generalization of One2Set whose results nearly drop by half for non-computer science domains, and that is even surpassed by MultiPartiteRank.
This latter delivers consistent, albeit modest, performance across domains which makes it relevant as an estimator of lower-bound performance for research on domain adaptation.


\subsection{Confidence Ranking of synthetic samples}
\label{subsec:confidence-ranking}

The purpose of \silk{} is to generate small, high quality in-domain data for fine-tuning keyphrase generation models.
Accordingly, synthetic samples are ordered by confidence (described in \S\ref{subsec:step-3}) and only the top-$N$ ranked samples are employed for adapting models. 
To validate the quality of our ranking, and consequently the effectiveness of our keyphrase candidate scoring function (see Equation~\ref{eq:scoring}), we compare the performance of BART-FT when we continue the fine-tuning with the top-1K, bottom-1K and a random selection of 1K samples.
Results are presented in Table~\ref{tab:ordering}.
We note that, uniformly across the three domains, the random and top-1K ordering schemes lead to improvements, with top-1K yielding the best results.
In contrast, using the least confident samples (bottom-1K) systematically degrades the performance.
Insights from these results are twofold: 1)~our confidence ranking proves to be beneficial for selecting high-quality synthetic samples, and 2)~even samples beyond the top-1K are qualitative enough for domain adaptation.

\begin{table}[hb!]
    \centering
    \resizebox{\linewidth}{!}{%
    \input{tables/ordering}
    }
    \caption{Performance of BART-FT fine-tuned on the top-1K, bottom-1K and random-1K (averaged over 5 runs with different seed values) samples. $\dagger$ indicates significance over BART-FT.} 
    \label{tab:ordering}
\end{table}

\subsection{Forgetting of Domain Adaptation}
\label{subsec:forgetting}

Although continued training is effective for domain adaptation, it has been found to adversely affect performance in the initial domain for language generation tasks~\cite{li-etal-2022-overcoming}.
Here, we investigate whether this phenomenon, referred to in the literature as catastrophic forgetting~\cite{FRENCH1999128}, also manifests in our adapted models.
Table~\ref{tab:cross-domains} presents the results of our domain adapted BART-FT models (using 1K synthetic samples) on the KP20k test set.
Overall, we observe no drop in performance for our adapted models.
Rather surprisingly, we notice small improvements in $F_1@k$ scores over the initial BART-FT model.
Upon closer examination, these gains derive from improved extractive capabilities, while the scores for absent keyphrases consistently degrade.
We hypothesise that the domain adaption process makes the model lose generative ability and reinforces its extractive capability which translates more effectively across domains.

\begin{table}[hb!]
    \centering
    \resizebox{\linewidth}{!}{%
    \input{tables/cross-domains}

    }
    \caption{Performance comparison of BART-FT and its adaptions (\silk{} 1K) on the KP20k test set.}
    \label{tab:cross-domains}
\end{table}

\subsection{Qualitative analysis}
\label{subsec:manual-evaluation}

We further examine the quality of the synthetic samples produced with \silk{} by conducting a manual evaluation of the top-100 samples of the \texttt{nlp} domain.\footnote{Annotation guidelines and examples can be found in~\ref{subsec:appendices-human-evaluation}.}
Annotators were instructed to assess the relevance of silver-standard keyphrases using a 3-point scale: ``\textit{not relevant}'', ``\textit{partially relevant}'' and ``\textit{relevant}''.
Additionally, we requested annotators to assess the well-formedness of the keyphrases with a binary rating.
To quantify the qualitative difference between \silk{} keyphrases and automatically generated ones, we perform a second round of human evaluation for BART-FT utilizing the same top-100 samples.
Table~\ref{tab:manual-eval} presents the results of our qualitative analysis.
First, we note that nearly all \silk{} keyphrases are well-formed, with any exceptions attributable to tagging errors (e.g.~``\textit{inter alia}'').
More importantly, we observe that 80\% of \silk{} keyphrases are relevant, demonstrating the effectiveness of our method.
In contrast, only 54.5\% of the keyphrases generated by BART-FT are deemed relevant, which explains why the self-learning approach to domain adaptation falls short.
We also note that BART-FT tends to generate more keyphrases ($\approx$5.5 per doc.), many of which are broader terms that are often irrelevant for the NLP domain (e.g.~``\textit{natural language processing}'', ``\textit{statistics}'' or ``\textit{machine learning}'').

\begin{table}[htb!]
    \centering
    \resizebox{\linewidth}{!}{%
    \input{tables/manual-eval}
    }
    \caption{Human evaluation results (\%) in terms of well-formedness and relevance of the top-100 \texttt{nlp} samples generated by \silk{} and re-annotated using BART-FT. 
    }
    \label{tab:manual-eval}
\end{table}

\subsection{Bias towards Highly Cited Papers}
\label{subsec:bias}

Since our method leverages citation contexts, it produces synthetic samples that are inherently biased towards highly cited papers and their corresponding keyphrases.
To investigate whether this bias is present in the adapted BART-FT models, we measure how frequently they generate keyphrases found in the synthetic samples and compare this number to that of our initial model.
Results are presented in Table~\ref{tab:bias}.
We observe only minor differences in the number of generated keyphrases from the synthetic samples, suggesting no apparent bias.
Conversely, we notice that the adapted models produce fewer of these keyphrases, as evidenced by the negative scores.
We attribute this to the few-shot fine-tuning, which may not sufficiently affect the model weights to propagate bias, and reinforces the extractive capabilities of the models, thereby making them less sensitive to bias.

\begin{table}[htb!]
    \centering

\input{tables/bias}
    \caption{Difference in the number of generated keyphrases found in \silk{} samples between BART-FT and its adaptations; a negative number means the adapted model generates fewer keyphrases from highly cited papers.}
    \label{tab:bias}
\end{table}

\section{Conclusion and Future Work}

In this paper, we propose \silk{}, an unsupervised method that relies on citation contexts to create silver-standard data for adapting keyphrase generation models to new domains.
We conduct experiments across three distinct scientific domains and demonstrate the effectiveness of our method for domain adaptation by few-shot fine-tuning a pre-trained model for keyphrase generation.
Our results show significant improvements in in-domain performance with 1K synthetic samples over strong baselines and self-supervised domain adaptation.
We further validate the quality of the synthetic samples created by \silk{} through human evaluation and analysis.

Our work addresses the issue of domain adaptation in keyphrase generation by introducing a solution that leverages citation contexts.
Considering that citing papers is the \textit{de-facto} means for discussing past work in scientific writing, we argue that it is possible to generate silver-standard data for most domains, provided that there is a minimal number of papers available.
Such data would not only be useful for adapting existing models to new domains but also for keeping them up-to-date, given the rapid expansion of scientific literature and the evolving terminology across all domains.

\section*{Limitations}

While our proposed method is both straightforward and effective, it is important to acknowledge its limitations.
First, we did not optimize each component of our method, relying instead on heuristics for selecting and filtering citation contexts and scoring silver-standard keyphrases using a simple combination of criteria.
Since our work focuses on generating synthetic data for domain adaptation, and we did not search for the optimal fine-tuning parameters, and also relied on a single pre-trained model (BART-base).
Even though we have conducted extensive experiments across three domains, it remains unclear how our findings generalize to other or larger pre-trained models.
Manually evaluating the quality of automatically generated keyphrases is inherently subjective.
Although we developed simple and detailed guidelines to minimize variability in assessments, it remains unclear how the results from our qualitative analysis extend beyond the top 100 samples in \texttt{nlp} or to the other two domains.

\section*{Acknowledgements}

This work was supported by the French National Research Agency (ANR) through the DELICES project (ANR-19-CE38-0005-01), and by the Defense Innovation Agency (AID) and the National Centre for Scientific Research (CNRS) through the NaviTerm project (convention 2022 65 0079 CNRS Occitanie Ouest).

\bibliography{anthology-small,custom}

\clearpage

\appendix

\section{Appendices}
\label{sec:appendix}

\subsection{Related work}

Keyphrase generation was first introduced by~\cite{liu-etal-2011-automatic} and subsequently formulated as a sequence-to-sequence language generation task by~\cite{meng-etal-2017-deep}.
They proposed an RNN-based encoder-decoder model with attention and copy mechanisms, which was later enhanced by the addition of decoding constraints to improve keyphrase diversity~\cite{chen-etal-2018-keyphrase,zhao-zhang-2019-incorporating,bahuleyan-el-asri-2020-diverse,yuan-etal-2020-one,Huang_Xu_Jiao_Zu_Zhang_2021}, or by learning to encode the structural information of input documents~\cite{ye-wang-2018-semi,10.1609/aaai.v33i01.33016268,kim-etal-2021-structure}. 
Later work switched to Transformers-based models~\cite{meng-etal-2021-empirical,ye-etal-2021-one2set,ahmad-etal-2021-select}, reporting better performance.
Recently, pre-trained language models (PLMs) have been used for keyphrase generation, predominantly through continued fine-tuning~\cite{zhao-etal-2022-keyphrase,meng-etal-2023-general,wu-etal-2023-rethinking-model}.

Our work also intersects with unsupervised models for keyphrase generation~\cite{shen2022unsupervised,do-etal-2023-unsupervised}, which evaluate the informativeness of keyphrases based on their semantic similarity to the source document.
Another direction to mitigate the data scarcity issue in keyphrase generation involves leveraging both labeled and unlabeled data for training.
\citet{ye-wang-2018-semi} proposed a self-learning approach to augment the training data with synthetic samples.
Similarly, \citet{meng-etal-2023-general} extended this concept to adapt models to new domains by generating domain-specific synthetic samples.
In a low-resource setting, \citet{garg-etal-2023-data} introduced a data augmentation method that leverages the full text of the documents to add diversity to the training samples.

Our work is closely related to the use of citation contexts in automated models for producing keyphrases.
For keyphrase extraction, \citet{das_gollapalli2014extracting} proposed a graph-ranking approach that leverages citation contexts while scoring candidates, and \citet{caragea-etal-2014-citation} use the occurrence of candidates in citation contexts as a feature in a supervised model.
For keyphrase generation, \citet{garg-etal-2022-keyphrase} proposed to append citation contexts to enrich the input document.



\subsection{Implementation Details}
\label{subsec:implementation-details}

We use the BART-base model weights as our initial pre-trained language model and perform fine-tuning
on the KP20k training set for 15 epochs.
We use the AdamW optimizer with a learning rate of 1e-5 and a batch size of 24.
Fine-tuning the model using 2 Nvidia GeForce RTX 2080 took 62 hours.

For adapting BART-FT to a each domain, we continue fine-tuning on $N \in \{500, 1K, 2K\}$ synthetic samples for 3 epochs.
We use the AdamW optimizer with a learning rate of 1e-6 and a batch size of 16.
Few-shot fine-tuning, conducted on a MacBook Pro M1 Max, required an average of 5 minutes per model, totaling 3 hours for all 12 models per domain.

For MultiPartiteRank, we use the author's implementation provided by the \texttt{pke} toolkit.\footnote{\url{https://github.com/boudinfl/pke}}
For Yake, we use the author's implementation.\footnote{\url{https://github.com/LIAAD/yake}}
For KeyBART, we use the model weights released by the authors and the suggested parameter settings (i.e.~beam width = 50, maximum generated sequence length = 40 tokens).\footnote{\url{https://huggingface.co/bloomberg/KeyBART}}
For One2Set, we use the model weights released by the authors.\footnote{\url{https://github.com/jiacheng-ye/kg_one2set}}

\subsection{Guidelines for manual evaluation}
\label{subsec:appendices-human-evaluation}

We evaluate the silver-standard keyphrases created by \silk{} and those generated by BART-FT along two criteria: their relevance with respect to the source document, and their well-formedness.
Annotators (authors of this paper) were given the title, the abstract and access to the full-text paper when evaluating the quality of the keyphrases.
We perform manual evaluation on the top-100 synthetic samples generated by \silk{}, confined to the \texttt{nlp} domain for which annotators have expertise.
\begin{enumerate}[wide, itemsep=0pt, leftmargin =*]
  \item[\textbf{Relevance}] is assessed on a 3-point scale, where 0 indicates that the keyphrase is not relevant, 1 that it is partially relevant (i.e.~covering a related concept) and 2 that it is relevant to the source document.
  \item[\textbf{Well-formedness}] is assessed on a binary scale, with 0 indicating that the keyphrase lacks proper form, such as being improperly structured (e.g.~``\textit{algorithms and data structures}'') or not forming a self-contained phrase (e.g.~``\textit{large amount}''), while 1 signifies that the keyphrase is well-formed.
\end{enumerate}

Orthographic variants occurring in a set of keyphrases (e.g.~``\textit{co-reference resolution}'' and ``\textit{coreference resolution}'') are identified, and only one of them is considered as relevant.
We do not consider abbreviations as variants of their expanded forms.
Broader terms such as ``\textit{natural language processing}'' or ``\textit{neural networks}'' are generally considered as too generic and not relevant.

An example of output for \silk{} and BART-FT is shown is Table~\ref{tab:silk_example}.

\begin{table*}[b]
    \centering
    \small
    \resizebox{\textwidth}{!}{%
    \input{tables/silk_example}

    }
    \caption{Examples of document (title and abstract) from the \texttt{nlp} domain with silver-standard keyphrases generated by \silk{} and automatically generated keyphrases from BART-FT.}
    \label{tab:silk_example}
\end{table*}




\clearpage

\subsection{Sources used for collecting test data}
\label{subsec:appendic-venues-for-gold}

\begin{table}[!htb]
    \centering
    \footnotesize
    \resizebox{.5\textwidth}{!}{
    \input{tables/nlp-gold-venues}
    }
    \caption{Detailed information on the sources of the test documents for the \texttt{nlp} domain. \faHandPaper[regular] indicates that we manually selected the documents to filter out out-of-domain ones.}
    \label{tab:nlp-gold-venues}
\end{table}

\begin{table}[!htb]
    \centering
    \resizebox{.5\textwidth}{!}{
    \input{tables/astro-gold-venues}
    }
    \caption{Detailed information on the sources of the test documents for the \texttt{astro} domain.}
    \label{tab:astro-gold-venues}
\end{table}

\begin{table}[!htb]
    \centering
    \resizebox{.5\textwidth}{!}{
    \input{tables/paleo-gold-venues}
    }
    \caption{Detailed information on the sources of the test documents for the \texttt{paleo} domain.}
    \label{tab:paleo-gold-venues}
\end{table}

\begin{table*}[!htb]
    \centering
    \resizebox{\textwidth}{!}{%
    \input{tables/paleo-sources}
    }
    \caption{Detailed information on the sources of the scientific papers collected for the Paleontology corpus.}
    \label{tab:paleo-sources}
\end{table*}

\begin{table*}[t]
    \centering
    \resizebox{\textwidth}{!}{%
    \input{tables/pres-abs-results}

    }
    \caption{Performance of keyphrase generation models on the \texttt{nlp}, \texttt{astro} and \texttt{paleo} domains for present and absent keyphrases separately. Values in \textbf{bold} indicate best scores and $\dagger$ indicates significance over BART-FT.}
    \label{tab:pres-abs-results}
\end{table*}

\end{document}

%% file: tables/data_statistics.tex

         

\begin{tabular}{r|rrr}
\toprule
  ~ & \textbf{\texttt{nlp}} & \textbf{\texttt{astro}} & \textbf{\texttt{paleo}} \\
\midrule
    \# documents & 65\,662 & 198\,349 & 12\,353 \\
    \# citation contexts & 260\,324  & 133\,320 & 53\,133 \\
    \# cited doc & 32\,448 & 20\,436 & 3\,252 \\
    cites / doc & 6.0  & 3.2 & 1.6 \\
    phrases / cited doc & 72.9 & 76.8 & 87.4 \\
\midrule
\multicolumn{4}{c}{\diagbox[height=\line,width=.5\textwidth,font=\footnotesize\itshape\bfseries]{\raisebox{-.15cm}{$\downarrow$ \texttt{silk} (top-1K - {\scriptsize all})}}{\raisebox{.15cm}{datasets $\uparrow$}}} \\

\midrule


    
    doc len. (tokens) & 149 & 202 & 278 \\ 
    

    keyphrase / doc & 3.9 {\scriptsize 3.6} 
                    & 3.6 {\scriptsize 3.5}
                    & 3.6 {\scriptsize 3.6} \\
    
    keyphrase len. & 1.8 & 1.9 & 1.6 \\ 


    \% abs keyphrases & 23.7 {\scriptsize  21.8}
                      & 16.7 {\scriptsize  14.8}
                      & 4.3 {\scriptsize  5.3}\\
    
\bottomrule
\end{tabular}


%% file: tables/test_data.tex
\begin{tabular}{r|rr|rrrrr}
    \toprule
        \textbf{Dataset} &
        \textbf{\#doc} &
        \textbf{len\textsubscript{doc}} &
        \textbf{\#kp} &
        \textbf{len\textsubscript{kp}} &
        \textbf{\%abs} \\
    \midrule
        KP20k   & 20K & 176 & 5.2 & 2.0 & 41.5 \\ 
    \midrule
        \texttt{nlp}   & 212 & 210 & 4.1 & 2.0 & 36.7 \\
        \texttt{astro} & 255 & 224 & 4.9 & 2.1 & 47.8 \\
        \texttt{paleo} & 244 & 255 & 5.5 & 1.5 & 38.6 \\
    \midrule
        Inspec  & 500 & 134 &  9.8 & 2.3 & 21.4 \\
        NUS     & 211 & 182 & 11.7 & 2.1 & 45.2 \\
        SemEval & 100 & 203 & 14.5 & 2.1 & 60.7 \\
    \bottomrule
\end{tabular}

%% file: tables/baseline-results.tex
\newcolumntype{R}{>{\scriptsize}r}

\begin{tabular}{l| r@{\hspace{.5em}}R@{\hspace{.4em}}R r@{\hspace{.5em}}R@{\hspace{.4em}}R r@{\hspace{.5em}}R@{\hspace{.4em}}R}
\toprule
    \multirow{2}{*}{\textbf{Model}} & 
    \multicolumn{3}{c}{$F_1@M$} & 
    \multicolumn{3}{c}{$F_1@5$} & 
    \multicolumn{3}{c}{$F_1@10$} \\
    \cmidrule(lr){2-4}
    \cmidrule(lr){5-7}
    \cmidrule(lr){8-10}
    ~ &
    all & pres & abs &
    all & pres & abs &
    all & pres & abs \\
\midrule
    MPRank  & 
            \multicolumn{3}{c}{-} & 
            14.8 & 18.7 & - &
            13.7 & 16.2 & - \\

    YAKE  & 
            \multicolumn{3}{c}{-} & 
            14.5 & 18.5 & - &
            14.6 & 17.4 & - \\

    KeyBART &
        11.4 & 16.4 & 1.6 &
        11.9 & 17.4 & 1.9 &
        11.0 & 15.3 & 1.7 \\
    
    One2Set & 23.2 & 35.1 & \ssign{5.5} &
            23.5 & 29.9 & 4.2 &
            \multicolumn{3}{c}{-}\\

    BART-FT    & 
            \ssign{28.7} & \ssign{37.3} & 2.4 &
            \ssign{28.0} & \ssign{35.5} & \ssign{5.9} & 
            \ssign{25.4} & \ssign{29.2} & 5.8 \\
\bottomrule
\end{tabular}





%% file: tables/adaptation-results.v4.tex
\begin{tabular}{l@{\hspace{.5em}}c|rrr|rrr|rrr}
\toprule
    \multirow{2}{3em}{\textbf{Model}} &
    \multirow{2}{*}{\textbf{FT}} &
    \multicolumn{3}{c}{\textbf{\texttt{nlp}}} &
    \multicolumn{3}{c}{\textbf{\texttt{astro}}} &
    \multicolumn{3}{c}{\textbf{\texttt{paleo}}} \\
    \cmidrule(lr){3-5}
    \cmidrule(lr){6-8}
    \cmidrule(lr){9-11}
    ~ & ~ & 
    \small $F_1@M$ & \small $F_1@5$ & \small $F_1@10$ & 
    \small $F_1@M$ & \small $F_1@5$ & \small $F_1@10$ & 
    \small $F_1@M$ & \small $F_1@5$ & \small $F_1@10$ \\
\midrule
    MPRank &  & 
    - & 17.1 & 14.3 & 
    - & 13.4 & 11.7 &
    - & 13.5 & 13.8 \\
    YAKE &  & 
    - & 20.3 & 18.1 & 
    - & 11.5 & 11.8 &
    - &  9.6 & 11.3 \\
    KeyBART & &
    11.8 & 12.8 & 11.0  &
    11.8 & 11.4 & 10.6 &
    8.2 & 8.0 & 8.8 \\
    One2Set & ~ & 
    21.6 & 24.1 & - & 
    13.2  & 12.7 & - &
    13.4 & 12.1 & - \\
\midrule
    BART-FT & ~ & 
    30.8 & 29.8 & 24.9 & 
    19.2 & 18.6 & 16.6 &
    18.4 & 18.9 & 18.8 \\
\cmidrule{1-11}
    +self-learning &
    500 & 
    31.2 & 30.0 & 24.5 &
    19.2 & 19.0 & 16.2 &
    18.9 & 18.7 & 18.6 \\
    ~ & 1K & 
    30.7 & 30.7 & 24.1 &
    19.7 & 19.6 & 16.6 &
    19.5 & 19.2 & 18.8 \\
    ~ & 2K & 
    30.0 & 29.2 & 24.4 & 
    18.4 & 19.2 & 16.4 &
    19.2 & 19.7 & 18.4 \\
\cmidrule{1-11}
    +\silk{} (ours) &
    500 & 
    31.0 & 29.7 & \textbf{25.2} &
    18.9 & 19.3 & 17.1  &
    19.0 & 19.3 & 19.2 \\
    ~ & 1K & 
    \sign{33.7} & \sign{32.2} & \textbf{25.2} & 
    19.3 & \ssign{20.5} & \ssign{17.7} &
    \textbf{19.6} & \sign{20.4} & \textbf{19.5} \\
    ~ & 2K & 
    31.0 & 31.0 &  24.3 & 
    \textbf{20.4} & \sign{21.5} & \sign{17.9} &
    17.7 & 19.1 & 18.0 \\
\bottomrule
\end{tabular}

%% file: tables/ordering.tex
\newcolumntype{R}{>{\footnotesize}r}

\begin{tabular}{l| r@{\hspace{.3em}}R r@{\hspace{.3em}}R r@{\hspace{.3em}}R}
\toprule
    \multirow{2}{*}{\textbf{Model}} & 
    \multicolumn{2}{c}{\textbf{\texttt{nlp}}} & 
    \multicolumn{2}{c}{\textbf{\texttt{astro}}} & 
    \multicolumn{2}{c}{\textbf{\texttt{paleo}}} \\
    \cmidrule(lr){2-3}
    \cmidrule(lr){4-5}
    \cmidrule(lr){6-7}
    ~ &
    \tiny $F_1@M$ & \tiny $F_1@10$ &
    \tiny $F_1@M$ & \tiny $F_1@10$ &
    \tiny $F_1@M$ & \tiny $F_1@10$ \\
\midrule
    BART-FT & 
        30.8 & 24.9 & 
        19.2 & 16.6 & 
        18.4 & 18.8 \\ 
\cmidrule{1-7}
    +\silk{} (top) &
        \ssign{33.7} & 25.2 & 
        19.2 & \ssign{17.7} & 
        19.4 & 19.5 \\ 

    +\silk{} (ran) &
        \ssign{33.2} & 25.1 & 
         19.3 & 17.4 & 
         19.4 & 19.4\\ 
        
    +\silk{} (bot) &
        27.8 & 21.4 & 
        16.5 & 15.1 & 
        16.9 & 17.3 \\ 

\bottomrule
\end{tabular}

%% file: tables/cross-domains.tex
\newcolumntype{R}{>{\scriptsize}r}

\begin{tabular}{l| r@{\hspace{.5em}}R@{\hspace{.3em}}R r@{\hspace{.5em}}R@{\hspace{.3em}}R r@{\hspace{.5em}}R@{\hspace{.3em}}R}
\toprule
    \multirow{2}{*}{\textbf{Model}} & 
    \multicolumn{3}{c}{$F_1@M$} & 
    \multicolumn{3}{c}{$F_1@5$} & 
    \multicolumn{3}{c}{$F_1@10$} \\
    \cmidrule(lr){2-4}
    \cmidrule(lr){5-7}
    \cmidrule(lr){8-10}
    ~ &
    all & pres & abs &
    all & pres & abs &
    all & pres & abs \\
\midrule
    BART-FT &
        28.7 & 37.3 & 2.4 &
        28.0 & 35.5 & 5.9 &
        25.4 & 29.2 & 5.8 \\
\cmidrule{1-10}

    +\silk{}~(\texttt{nlp}) & 
    28.6 & 37.5 & 1.6 &
    28.3 & 35.9 & 5.5 &
    25.7 & 29.7 & 5.4 \\
    
    +\silk{}~(\texttt{astro}) & 
    28.7 & 37.8 & 1.7 & 
    28.7 & 36.4 & 5.9 &
    25.9 & 29.8 & 5.9 \\

    +\silk{}~(\texttt{paleo}) &
    28.4 & 37.5 & 1.4 &
    28.6 & 36.2 & 5.7 &
    25.8 & 29.7 & 5.6 \\
\bottomrule
\end{tabular}

%% file: tables/manual-eval.tex
\begin{tabular}{l|r|rrrrrr}
\toprule
    \multirow{2}{*}{\textbf{Model}} &  
    \multirow{2}{*}{\textbf{\#kp}} & 
    \multicolumn{2}{c}{\textbf{WFness}} &
    \multicolumn{3}{c}{\textbf{Relevance}}
    \\
    \cmidrule(lr){3-4}
    \cmidrule(lr){5-7}
    ~ & ~ & \small no & \small yes & \small no & \small part. & \small yes \\
\midrule
BART-FT  & 545 & 6.2 & 93.8 & 35.0 & 10.5 & 54.5 \\
\silk{}  & 411 & 2.9 & 97.1 & 11.9 &  8.0 & 80.0  \\

\bottomrule
\end{tabular}




%% file: tables/bias.tex
\begin{tabular}{l| rrr}
\toprule
    \textbf{Model} & 
    \textbf{\texttt{nlp}} & 
    \textbf{\texttt{astro}} &
    \textbf{\texttt{paleo}} \\
\midrule

    +\silk{} (500) & -0.04 & -0.02 & -0.02 \\
    +\silk{} (1K)  & -0.19 & -0.01 & -0.12 \\
    +\silk{} (2K)  & -0.37 & +0.18 & -0.23 \\

\bottomrule
\end{tabular}

%% file: tables/silk_example.tex
\begin{tabular}{ll}
\toprule
\multicolumn{2}{p{\textwidth}}{
    \textbf{Get To The Point: Summarization with Pointer-Generator Networks} (Bibkey: see-etal-2017-get)
} \\
\midrule
\multicolumn{2}{p{\textwidth}}{
    Neural sequence-to-sequence models have provided a viable new approach for abstractive text summarization (meaning they are not restricted to simply selecting and rearranging passages from the original text). However, these models have two shortcomings: they are liable to reproduce factual details inaccurately, and they tend to repeat themselves. In this work we propose a novel architecture that augments the standard sequence-to-sequence attentional model in two orthogonal ways. First, we use a hybrid pointer-generator network that can copy words from the source text via pointing, which aids accurate reproduction of information, while retaining the ability to produce novel words through the generator. Second, we use coverage to keep track of what has been summarized, which discourages repetition. We apply our model to the CNN / Daily Mail summarization task, outperforming the current abstractive state-of-the-art by at least 2 ROUGE points.
}\\
\cmidrule{1-2}
\silk{} & summarization, pointer-generator network, sequence-to-sequence model, copy mechanism, coverage mechanism \\
~ & \scriptsize{well-formedness: 1 1 1 1 1} \quad \scriptsize{relevance: 1 1 1 1 1} \\[.3em]
BART-FT & summarization, sequence-to-sequence models, attentional models, cnn,  daily mail, neural networks, text mining \\
~ & \scriptsize{well-formedness: 1 1 1 1 1 1 1} \quad \scriptsize{relevance: 1 1 1 1 1 0 0} \\
\bottomrule

 \\
 
\toprule
\multicolumn{2}{p{\textwidth}}{
    \textbf{Improving Neural Machine Translation Models with Monolingual Data} (Bibkey: sennrich-etal-2016-improving)
} \\
\midrule
\multicolumn{2}{p{\textwidth}}{
    Neural Machine Translation (NMT) has obtained state-of-the art performance for several language pairs, while only using parallel data for training. Target-side monolingual data plays an important role in boosting fluency for phrase-based statistical machine translation, and we investigate the use of monolingual data for NMT. In contrast to previous work, which combines NMT models with separately trained language models, we note that encoder-decoder NMT architectures already have the capacity to learn the same information as a language model, and we explore strategies to train with monolingual data without changing the neural network architecture. By pairing monolingual training data with an automatic back-translation, we can treat it as additional parallel training data, and we obtain substantial improvements on the WMT 15 task English German (+2.8-3.7 BLEU), and for the low-resourced IWSLT 14 task Turkish->English (+2.1-3.4 BLEU), obtaining new state-of-the-art results. We also show that fine-tuning on in-domain monolingual and parallel data gives substantial improvements for the IWSLT 15 task English->German.
}\\
\cmidrule{1-2}
\silk{} & neural machine translation, monolingual data, back-translation, data augmentation, synthetic parallel corpus\\
 ~ & \scriptsize{well-formedness: 1 1 1 1 1} \quad \scriptsize{relevance: 1 1 1 1 1} \\[.3em]
BART-FT & neural machine translation, monolingual data, language models, back-translation, language modeling and \\
& translation, parallel training data \\
~ & \scriptsize{well-formedness: 1	1	1	1	0	1} \quad \scriptsize{relevance: 1	1	1	1	0	0.5} \\[.3em]
\bottomrule
\end{tabular}

%% file: tables/nlp-gold-venues.tex
\begin{tabular}{r|lrr}
\toprule
    \textbf{Source}  & \textbf{Session / Volume} &  \textbf{\#nb} \\
\midrule
    SIGIR 2023 & Language Models & 6 \\
               & Question Answering & 3 \\
               & Summarization \& Text Generation & 5 \\
               & Short Research Papers\textsuperscript{\faHandPaper[regular]} & 16 \\
    CIKM 2023  & Natural Language & 24 \\
    WSDM 2023  & Language Models and Text Mining & 6 \\
    SIGIR 2022 & NLP and Semantics & 8 \\
               & Question Answering & 4 \\
               & Sentiment Analysis and Classification & 5 \\
               & Short Research Papers\textsuperscript{\faHandPaper[regular]} & 14 \\
    CHI 2022   & Natural Language & 5 \\
    LREC 2022  & Oral sessions\textsuperscript{\faHandPaper[regular]} & 81 \\
    NLP Journal\footnote{\url{https://www.sciencedirect.com/journal/natural-language-processing-journal}} & Volumes 2-5 & 35 \\
\midrule
    & Total & 212 \\
\bottomrule
\end{tabular}

%% file: tables/astro-gold-venues.tex
\begin{tabular}{r|lr}
\toprule
    \textbf{Source}  & \textbf{Category / Year} & \textbf{\#nb} \\
\midrule
    arXiv                              & astro-ph.HE (High Energy Astro. Phenomena)& 20 \\
    (oct$\rightarrow$dec 2023)       & astro-ph.CO (Cosmology and Nongalactic Astro.) & 20 \\
    ~                                  & astro-ph.IM (Instrumentation and Methods for Astro.) & 20 \\
    ~                                  & astro-ph.SR (Solar and Stellar Astro.) & 20 \\
    ~                                  & astro-ph.EP (Earth and Planetary Astro.) & 20 \\
    ~                                  & astro-ph.GA (Astro. of Galaxies) & 20 \\
    Frontiers in Astro. &  2022-23 (selected using arXiv keywords) & 76 \\ 
    Astrophysics & 2022-23 & 59 \\
\midrule
    & Total & 255 \\
\bottomrule
\end{tabular}

%% file: tables/paleo-gold-venues.tex
\begin{tabular}{r|lrr}
\toprule
    \textbf{Source}  & \textbf{Year} & \textbf{\#nb} \\
\midrule
    Palaeontologia Electronica    & 2023-24 & 21\\
    Acta Palaeontologica Polonica & 2023 & 22 \\
    Palaeontology                 & 2023 & 26 \\
    Cretaceous Research           & 2024 & 20 \\
    Palaeogeography, Palaeoclimatology, Palaeoecology & 2024 & 47 \\
    Papers in Palaeontology & 2023 & 29 \\
    Proc. Royal Soc. B: Biological Sciences & 2023 & 25\\
    Biology Letters & 2023 & 25 \\
    Palaeobiodiversity and Palaeoenvironments & 2023 & 16 \\
\midrule
    & Total & 244 \\
\bottomrule
\end{tabular}

%% file: tables/paleo-sources.tex

\begin{tabular}{lccr}
\toprule
\textbf{Source} & \textbf{Licence} & \textbf{Year} & \textbf{\#nb} \\
\midrule
Acta Geologica Sinica   & open & 2021-2023 & 21 \\
Acta Palaeontologica Polonica   & open & 2002-2023 & 1\,398 \\
Alcheringa: An Australasian Journal of Palaeontology   & open & 2016-2023  & 32 \\
Carnets de Geologie & open & 2015-2023 & 147 \\
Cretacious Research   & open & 2019-2023 & 81 \\
Journal of paleontology  & open & 2015-2023 & 144 \\
Journal of Systematic Palaeontology   & open & 2016-2023  & 34 \\
Journal of Vertebrate Paleontology  & open & 2013-2023  & 95 \\
Lethaia   & open/free & 2018-2023  & 239 \\
Nature & open & 2010-2023 & 705 \\
Palaeobiodiversity and Palaeoenvironments & open & 2002-2023 & 811 \\
Palaeodiversity  & open & 2016-2023 & 74 \\
Palaeogeography, Palaeoclimatology, Palaeoecology   & open & 2019-2023 & 201 \\
Palaeontologia Electronica   & open & 1998–2023 & 841 \\
Palaeontology  & open/free & 1999-2023  & 1\,474 \\
Paleobiology   & open & 2013-2023  & 133 \\
Paleoceanography and Paleoclimatology & open & 2014-2023 & 128 \\
PalZ & open/free & 2009-2023 & 651 \\
Papers in Palaeontology   & open & 2015-2023  & 71 \\
Plos Paleontology   & open & 2011-2017  & 237 \\
Proceedings of the Royal Society B: Biological Sciences  (paleontology)  & open/free & 2009-2023  & 354 \\
PubMedfreef ulltext (query="Paleontology[MeSH Terms]")  & open/free & 1955-2023 & 3\,462 \\
Research in Paleontology and Stratigraphy  & open & 2019-2023 & 157 \\
Royal Society   Science  (paleontology)  & open/free & 2014-2023 & 270 \\
Royal Society Biology Letters  (paleontology)  & open/free & 2009-2023 & 235 \\
Swiss Journal of Palaeontology & open/free & 2011-2023 & 282 \\
Trends in Ecology and Evolution  (paleobiology)  & open & 2020-2022  & 15 \\
Zookeys (paleontology) & open & 2015-2023 & 61 \\
\cmidrule{1-4}
& & Total & 12\,353\\
\bottomrule
\end{tabular}

%% file: tables/pres-abs-results.tex
\begin{tabular}{l@{\hspace{.5em}}c| rrrrrr | rrrrrr | rrrrrr}
\toprule
    \multirow{2}{3em}{\textbf{Model}} &
    \multirow{2}{*}{\textbf{FT}} &
    \multicolumn{6}{c}{\textbf{\texttt{nlp}}} &
    \multicolumn{6}{c}{\textbf{\texttt{astro}}} &
    \multicolumn{6}{c}{\textbf{\texttt{paleo}}} \\
    \cmidrule(lr){3-8}
    \cmidrule(lr){9-14}
    \cmidrule(lr){15-20}
    ~ & ~ & 
    \multicolumn{2}{c}{$F_1@M$} & \multicolumn{2}{c}{$F_1@5$} & \multicolumn{2}{c}{$F_1@10$} & 
    \multicolumn{2}{c}{$F_1@M$} & \multicolumn{2}{c}{$F_1@5$} & \multicolumn{2}{c}{$F_1@10$} & 
    \multicolumn{2}{c}{$F_1@M$} & \multicolumn{2}{c}{$F_1@5$} & \multicolumn{2}{c}{$F_1@10$} \\

    \cmidrule(lr){3-4} \cmidrule(lr){5-6} \cmidrule(lr){7-8}
    \cmidrule(lr){9-10} \cmidrule(lr){11-12} \cmidrule(lr){13-14}
    \cmidrule(lr){15-16} \cmidrule(lr){17-18} \cmidrule(lr){19-20}

    ~ & ~ & 
    \small pres & \small abs & \small pres & \small abs & \small pres & \small abs & 
    \small pres & \small abs & \small pres & \small abs & \small pres & \small abs & 
    \small pres & \small abs & \small pres & \small abs & \small pres & \small abs \\
    
\midrule

    MPRank &  & 
    \multicolumn{2}{c}{-} & 20.4 & - & 16.2 & - &
    \multicolumn{2}{c}{-} & 17.7 & - & 14.2 & - &
    \multicolumn{2}{c}{-} & 16.4 & - & 15.7 & - \\
    

    Yake &  & 
    \multicolumn{2}{c}{-} & 24.3 & - & 20.5 & - &
    \multicolumn{2}{c}{-} & 15.4 & - & 14.3 & - &
    \multicolumn{2}{c}{-} & 11.9 & - & 12.9 & - \\


    KeyBART &  & 
    16.6 & 0.8 & 17.3 & 1.5 & 14.0 & 1.5 &
    19.7 & \textbf{2.1} & 17.8 & \textbf{2.1} & 13.6 & \textbf{1.7} &
    11.6 & \textbf{1.8} & 12.5 & \textbf{1.9} & 13.0 & \textbf{1.7} \\


    One2Set & ~ & 
    36.1 & \textbf{3.3} & 28.3 & 2.8 & \multicolumn{2}{c|}{-} & 
    20.1 & 1.6 & 17.3 & 1.2 & \multicolumn{2}{c|}{-} &
    18.6 & 0.3 & 16.5 & 0.2 & \multicolumn{2}{c}{-}  \\


\midrule

    BART-FT & ~ & 
    38.8 & 2.0 & 36.8 & \textbf{3.7} & 27.9 & \textbf{3.6} & 
    26.7 & 0.2 & 25.6 & 1.4 & 20.2 & 1.4 &
    23.5 & 0.5 & 24.3 & 1.3 & 22.6 & 1.2
    \\



\cmidrule{1-20}

    ~+self-learning & 500 &  
    38.9 & 2.0 & 36.9 & 3.0 & 27.7 & 3.0 &
    26.5 & 0.0 & 26.0 & 0.7 & 19.9 & 0.7 &
    24.1 & 0.5 & 23.8 & 1.2 & 22.6 & 1.1 
    \\


    ~ & 1K &  
    38.2 & 2.0 & 36.8 & 3.1 & 27.6 & 3.0 &
    27.0 & 0.0 & 26.1 & 0.5 & 20.5 & 0.5 &
    24.5 & 0.2 & 24.8 & 0.6 & 22.9 & 0.7
    \\
    

    ~ & 2K &
    37.2 & 2.6 & 36.8 & 3.2 & 27.8 & 3.0 &
    25.3 & 0.2 & 25.3 & 0.5 & 20.2 & 0.7 &
    24.3 & 0.2 & 25.7 & 0.6 & 22.7 & 0.6 
    \\

\cmidrule{1-20}

    ~+\silk{} (ours) & 500 & 
    39.1 & 0.5 & 36.2 & 2.8 & \textbf{28.2} & 2.9 &
    26.4 & 0.2 & 26.7 & 1.1 & \ssign{21.1} & 1.1 &
    24.4 & 0.5 & 24.8 & 1.0 & \textbf{23.5} & 0.9 
    \\


    ~ & 1K &
    \sign{41.7} & 1.2 & \textbf{38.3} & 3.3 & 28.1 & 3.4 &
    27.0 & 0.0 & \ssign{27.7} & 1.1 & \ssign{21.7} & 1.0 &
    \textbf{25.4} & 0.7 & \sign{26.3} & 0.6 & \textbf{23.5} & 0.5
    \\


    ~ & 2K &
    38.8 & 0.3 & 37.2 & 2.8 & 27.3 & 2.8 &
    \textbf{29.0} & 0.0 & \sign{28.9} & 0.2 & \sign{22.2} & 0.2 & 
    23.2 & 0.0 & 24.5 & 0.5 & 21.9 & 0.5 
    \\


\bottomrule
\end{tabular}